\documentclass[conference]{IEEEtran}
\usepackage{cite}
\usepackage{amsmath,amssymb,amsfonts}
\usepackage{algorithmic}
\usepackage{graphicx}
\usepackage{textcomp}
\usepackage{xcolor}
\usepackage{mathtools, stmaryrd}
\usepackage{amssymb}
\usepackage{graphicx}
\usepackage{subcaption}
\usepackage{tabularx}
\usepackage{supertabular}
\usepackage{multirow}
\usepackage{boxedminipage}
\usepackage{eurosym}
\usepackage{enumitem}   
\usepackage[us]{datetime} 
\usepackage{amsbsy}
\usepackage{algorithm}
\usepackage{algorithmic}
\usepackage[draft=False]{hyperref} 
\hypersetup{
     colorlinks=true,
     linkcolor=blue,
     filecolor=blue,
     citecolor=blue,      
     urlcolor=blue,
     }

\newcommand{\startdataset}{\formatdate{04}{04}{2020}}
\newcommand{\lastdataset}{\formatdate{14}{09}{2020}}
\newcommand{\figuredatecomparison}{\formatdate{02}{08}{2020}}
\renewcommand{\arraystretch}{1.4} 

\def\BibTeX{{\rm B\kern-.05em{\sc i\kern-.025em b}\kern-.08em
    T\kern-.1667em\lower.7ex\hbox{E}\kern-.125emX}}
\begin{document}

\title{Deep learning-based multi-output quantile forecasting of PV generation}

\author{\IEEEauthorblockN{Jonathan Dumas, Colin Cointe, Xavier Fettweis, Bertrand Corn\'elusse}
\IEEEauthorblockA{\textit{Departments of computer science, electrical engineering, and geography} \\
\textit{University of Li\`ege, Belgium}\\
\{jdumas, xavier.fettweis, bertrand.cornelusse\}@uliege.be, colin.cointe@mines-paristech.fr}
}

\IEEEoverridecommandlockouts
\IEEEpubid{\makebox[\columnwidth]{978-1-6654-3597-0/21/\$31.00~\copyright2021 IEEE \hfill} \hspace{\columnsep}\makebox[\columnwidth]{ }}

\maketitle

\IEEEpubidadjcol

\begin{abstract}                   
This paper develops probabilistic PV forecasters by taking advantage of recent breakthroughs in deep learning. A tailored forecasting tool, named \textit{encoder-decoder}, is implemented to compute intraday multi-output PV quantiles forecasts to efficiently capture the time correlation. The models are trained using quantile regression, a non-parametric approach that assumes no prior knowledge of the probabilistic forecasting distribution. The case study is composed of PV production monitored on-site at the University of Li\`ege (ULi\`ege), Belgium. The weather forecasts from the regional climate model provided by the Laboratory of Climatology are used as inputs of the deep learning models. The forecast quality is quantitatively assessed by the continuous ranked probability and interval scores. The results indicate this architecture improves the forecast quality and is computationally efficient to be incorporated in an intraday decision-making tool for robust optimization.
\end{abstract}


\section{Introduction}

The Intergovernmental Panel on Climate Change special report\footnote{\url{https://www.ipcc.ch/sr15/}} on the impacts of global warming of 1.5\textdegree C above pre-industrial levels and related global greenhouse gas emission pathways’ presents several scenarios of decarbonisation of the electricity sector with a renewable share that should reach in 2030 the interquartile range $[47, 65]$ ($[69, 86]$ in 2050).
Therefore, the development of renewable generations, typically from wind and photovoltaic (PV) sources has been facilitated by policy makers. However, the intermittent and uncertain nature of these sources is challenging the traditional operation of electricity networks. Market players require reliable decision-making tools to deal with uncertainty based on forecasts of renewable generation. 

In contrast to \textit{point} predictions, \textit{probabilistic} forecasts aim at providing decision-makers with full information about potential future outcomes \cite{morales2013integrating}. The various types of probabilistic forecasts range from quantile to density forecasts, and through prediction intervals. This paper focuses on \textit{quantile} forecasts that provide probabilistic information about future renewable power generation, in the form of a threshold level associated with a probability \cite{morales2013integrating}. Quantile regression \cite{koenker1978regression} is one of the most famous non-parametric approaches. It does not assume the shape of the predictive distributions and is implemented with neural networks, linear regression, gradient boosting, or any other regression techniques.

The following papers have gained our attention in PV probabilistic forecasting.
At the Global Energy Forecasting Competition 2014 \cite{hong2016probabilistic} solar forecasts were to be expressed in the form of 99 quantiles with various nominal proportions between zero and one.
A systematic framework for generating PV probabilistic forecasts is developed by \cite{golestaneh2016very}. A non-parametric density forecasting method based on Extreme Learning Machine is adopted to avoid restrictive assumptions on the shape of the forecast densities. 
A combination of bidirectional Recurrent Neural Networks (RNNs) with Long Short-Term Memory (LSTM) resulting in Bidirectional LSTM (BLSTM) is proposed by \cite{toubeau2018deep}. It has the benefits of both long-range memory and bidirectional processing. The BLSTM is trained by minimizing the quantile loss to compute quantile forecasts of aggregated load, wind and PV generation, and electricity prices on a day-ahead basis. Finally, an innovative architecture, referred to as encoder-decoder (ED), is developed by \cite{bottieau2019very} to generate reliable predictions of the future system imbalance used for robust optimization.

In this study, the forecast quality of the models is evaluated. It corresponds to the ability of the forecasts to genuinely inform of future events by mimicking the characteristics of the processes involved. We follow the framework proposed by \cite{lauret2017probabilistic,lauret2019verification} for evaluating the quality of solar irradiance probabilistic forecasts based on visual diagnostic tools and a set of scoring rules. Overall they indicate that two main attributes, reliability and resolution, characterize the quality of quantile forecasts.

This work exploits recent breakthroughs in the field of data science by using advanced deep learning structures, such as the encoder-decoder architecture \cite{bottieau2019very}, and quality metrics \cite{pinson2007non,lauret2019verification} to develop a tailored deep learning-based multi-output quantile forecaster. The goal is to capture the time correlation between time periods and to use this forecaster as input of a robust optimization model. For instance, to address the energy management system of a grid-connected renewable generation plant coupled with a battery energy storage device \cite{dumas2020stochastic}. 

Overall, the contributions can be summarized as follows.
First, a deep learning-based multi-output quantile architecture is used to compute prediction intervals of PV generation on a day-ahead and intraday basis. Specifically, the goal is to implement an improved probabilistic intraday forecaster, the encoder-decoder, to benefit from the last PV generation observations. This architecture is compared to a feed-forward neural network that is used as the benchmark model.
Second, the weather forecasts of the MAR climate regional model \cite{fettweis2017reconstructions} are used. It allows to directly take into account the impact of the weather forecast updates generated every six hours.
Finally, a proper assessment of the quantile forecasts is conducted by using a $k$-fold cross-validation methodology and probabilistic metrics. It allows computing average scores over several testing sets and mitigating the dependency of the results to specific days of the dataset. 

The remainder of this paper is organized as follows. Section \ref{sec:pb_statement} provides the non-parametric quantile forecasting framework considered. Section \ref{sec:forecasting_techniques} presents the forecasting techniques used to compute the quantile forecasts. Section \ref{sec:scoring} details the different metrics used to evaluate the quality of PV quantile forecasts. Section \ref{sec:case_study} describes the case study and presents the results. Finally, Section \ref{sec:conclusion} summarizes the main findings and highlights ideas for further work. 

\section{Non-parametric quantile forecasting}\label{sec:pb_statement}

Let $y_t$ be the PV power generation measured at time $t$, which corresponds to a realization of the random variable $Y_t$. Then let $f_t$ and $F_t$ be the probability density function (PDF) and related cumulative distribution function (CDF) of $Y_t$, respectively. Following the definition of probabilistic forecasting of \cite{morales2013integrating}, a probabilistic forecast issued at time $t$ for time $t + k$ consists of a prediction of the PDF (or equivalently, the CDF) of $Y_{t +k}$, or of some summary features. Various types of probabilistic forecasts have been developed, from quantile to density forecasts, and through prediction intervals. This study focuses on quantile regression \cite{koenker1978regression}, which is the most widely used type of probabilistic forecasting method. It is a non-parametric approach with no restrictive assumption on the shape and features of the predictive distributions. Indeed, empirical investigations \cite{golestaneh2016very} showed that PV power forecast errors do not follow common, \textit{e.g.} Gaussian, Beta, distributions.
Following the definition of \cite{morales2013integrating}, a (model-based) forecast $ \hat{y}_{t+k|t}$ of PV power generation is an estimate of some of the characteristics of the stochastic process $Y_{t +k}$ given a model $g$, its estimated parameters $\hat{\Theta}_t$ and the information set $\Omega_t$ gathering all data and knowledge about the processes of interest up to time $t$.

\subsection{Point forecasting}

A point prediction $ \hat{y}_{t+k|t}$ is a single-valued issued at time $t$ for $t + k$. It corresponds to the conditional expectation of $Y_{t+k}$ given $g$, $\hat{\Theta}_t$, and the information set $\Omega_t$ 
\begin{equation}\label{eq:point_forecast_def}
\hat{y}_{t+k|t} = \mathop{\mathbb{E}} \big[Y_{t+k|t} | g, \Omega_t, \hat{\Theta}_t\big] .
\end{equation}
A multi-output point forecast computed at $t$ for $t+k_1$ to $t+k_T$ is the vector $ \big[\hat{y}_{t+k_1|t}, \cdots,\hat{y}_{t+k_T|t}\big]^\intercal $ of size $T$.

\subsection{Quantile forecasting}

A quantile forecast $\hat{y}^{(q)}_{t+k|t}$ with nominal level $q$ is an estimate, issued at time $t$ for time step $t+k$ of the quantile $y^{(q)}_{t+k|t}$ for the random variable $Y_{t+k|t}$ given a model $g$, its estimated parameters $\hat{\Theta}_t$ and the information set $\Omega_t$
\begin{equation}\label{eq:quantile_forecast_def}
P[Y_{t+k|t} \leq \hat{y}^{(q)}_{t+k|t} | g, \Omega_t, \hat{\Theta}_t]  = q,
\end{equation}
or equivalently $\hat{y}^{(q)}_{t+k|t}  = \hat{F}^{-1}_{t+k|t}(q)$, with $\hat{F}$ the estimated cumulative distribution function of the continuous random variable $Y$.
Finally, a multi-output quantile forecast computed at $t$ for $t+k_1$ to $t+k_T$ is the matrix $\hat{Z}_t$ of dimensions $T \times Q$

\begin{equation}\label{eq:multi_output_quantile_forecast}
\hat{Z}_t =  \begin{bmatrix}
\hat{y}^{(q=1)}_{t+k_1|t} &  \ldots &  \hat{y}^{(q=Q)}_{t+k_1|t}\\
\vdots & \vdots & \vdots \\
\hat{y}^{(q=1)}_{t+k_T|t} & \ldots & \hat{y}^{(q=Q)}_{t+k_T|t}
\end{bmatrix},
\end{equation}
with $Q$ quantiles per time period.

\subsection{Quantile loss function}

Quantile regression consists of estimating quantiles by applying asymmetric weights to the mean absolute error. Following \cite{koenker1978regression}, the quantile loss function is
\begin{equation}\label{eq:quantile_loss_def}
\rho_q(x, y) =  \begin{cases} q \times (x - y) &    x > y \\
(1-q) \times (y-x)  &  x \leq y \end{cases} .
\end{equation}
For a given time period $t$, $\rho_q$ is summed over the forecasting time periods $k_1 \leq k \leq k_T$ and quantiles $q \in \mathcal{Q}$ to compute multi-output quantile forecasts at $t$ for $t+k_1$ to $t+k_T$
\begin{equation}\label{eq:multioutput_quantile_loss_def}
L_t = \sum_{k_1 \leq k \leq k_T} \sum_{q \in \mathcal{Q}} \rho_q \big(y_{t+k}, \hat{y}^{(q)}_{t+k|t} \big)  .
\end{equation}
Finally, the model $g$ is trained by minimizing $\frac{1}{|\Omega_t|} \sum_{t' \in \Omega_t} L_{t'}$. Note in the case of perfect prediction, the quantile loss cannot be differentiated, and a smooth approximation of (\ref{eq:quantile_loss_def}) using the Huber norm \cite{bottieau2019very} is built.

\subsection{Prediction intervals (PIs)}

Prediction intervals (PIs) define the range of values within which the observation is expected to be with a certain probability, \textit{i.e.}, its nominal coverage rate \cite{pinson2007non}. 
Formally, a prediction interval $\hat{I}^{(\alpha)}_{t+k|t}$ issued at $t$ for $t + k$, defines a range of potential values for $Y_{t+k}$, for a certain level of probability $(1-\alpha)$, $\alpha \in [0,1]$. Its nominal coverage rate is
\begin{equation}\label{eq:PI_def1}
P\big[Y_{t+k} \in \hat{I}^{(\alpha)}_{t+k|t}| g, \Omega_t, \hat{\Theta}_t\big]  = 1 - \alpha.
\end{equation}
A central PI consists of centering the PI on the median where there is the same probability of risk below and above the median. A central PI with a coverage rate of $(1-\alpha)$ is estimated by using the quantiles $q=(\alpha/2)$ and $q=(1-\alpha/2)$. Its nominal coverage rate is
\begin{equation}\label{eq:PI_def}
\hat{I}^{(\alpha)}_{t+k|t}= \big[\hat{y}^{(q=\alpha/2)}_{t+k|t}, \hat{y}^{(q=1-\alpha/2)}_{t+k|t}\big].
\end{equation}
For instance, central PIs with a nominal coverage rate of 90 \%, \textit{i.e.}, $(1 - \alpha) = 0.9$, are defined by quantile forecasts with nominal levels of 5 and 95 \%.

\section{Forecasting techniques}\label{sec:forecasting_techniques}

\subsection{Gradient boosting regression (GBR)}

Gradient boosting builds an additive model in a forward stage-wise fashion \cite{hastie2009elements}. It allows for the optimization of arbitrary differentiable loss functions. In each stage, a regression tree is fit on the negative gradient of the given loss function.
The Scikit-learn \cite{scikit-learn} Python library is used to implement a gradient boosting regressor (GBR) with the quantile loss function. The learning rate is set to $10^{-2}$, the max depth to 5, and the number of estimators to 500. There is a GBR model per quantile as the library does not support multi-output quantile regression.

\subsection{Multilayer perceptron (MLP)}

A description of the most widely used "vanilla" neural network, the \textit{Multilayer perceptron} (MLP), is provided by \cite{hastie2009elements}. A MLP with a single hidden layer is considered for the day-ahead forecasts and as the benchmark for the intraday forecasts. MLPs with two and three hidden layers did not provide any significant improvement. The activation function is the Rectified Linear Unit (ReLU). The number of neurons of the hidden layer is $n_\text{input} + (n_\text{output} - n_\text{input}) / 2$, with $n_\text{input}$ and $n_\text{output}$ the number of neurons of the input and output layers, respectively. The learning rate is set to $10^{-2}$ and the number of epoch to 500 with a batch size of 8. It is implemented using the PyTorch Python library \cite{paszke2017automatic}. 

\subsection{Encoder-decoder (ED)}

Several technical information about recent advances in neural networks is provided by \cite{toubeau2018deep,bottieau2019very}. In particular, recurrent neural networks, have shown a high potential in processing and predicting complex time series with multi-scale dynamics. However, RNNs are known to struggle in accessing time dependencies more than a few time steps long due to the vanishing gradient problem. Indeed, back-propagated errors during the training stage either fades or blows up over time. Long Short-Term Memory and Gated Recurrent Units networks tackle this problem by using internal memory cells \cite{bottieau2019very}. A neural network composed of a LSTM and feed-forward layers, referred to as LSTM in the rest of the paper, is implemented for the day-ahead and intraday forecasts. The number of LSTM units is $n_\text{input} + (n_\text{output} - n_\text{input}) / 3$, and the number of neurons of the feed-forward layer $n_\text{input} + 2 \times (n_\text{output} - n_\text{input}) / 3$. 

An innovative architecture, referred to as encoder-decoder \cite{bottieau2019very}, is composed of two different networks and has recently shown promising results for translation tasks and speech recognition applications and imbalance price forecasting. The encoder-decoder processes features from the past, such as past PV observations, to extract the relevant historical information that is contained into a reduced vector of fixed dimensions, based on the last hidden state. Then, the decoder processes this representation along with the known future information such as weather forecasts.
A version of the encoder-decoder architecture (ED-1) is implemented with a LSTM as the encoder and a MLP as the decoder. In a second version (ED-2) the decoder is a LSTM followed by an additional feed-forward layer. Both versions of the encoder-decoder are used as intraday forecasters. In ED-1, the encoder has $2 \times n_\text{input}$ units with $n_\text{input}$ the number of neurons of the encoder input layer, features from the past. Then, the encoder output is merged with the weather forecasts becoming the decoder input layer that has $n_\text{output} /2$ neurons. In ED-2, the decoder has the same number of cells as the encoder, and the feed-forward layer is composed of $n_\text{output} /2$ neurons. The LSTM, ED-1, and ED-2 models are implemented using the TensorFlow Python library \cite{tensorflow2015-whitepaper}. The activation functions are the ReLU, the learning rate is set to $10^{-3}$, the number of epoch to 500 with a batch size of 64 for the three models.

A sensitivity analysis has been conducted to select the hyperparameters: number of hidden layers, neurons, epochs, and learning rate. Overall, increasing the number of hidden layers and neurons increases the model complexity. It can enhance the accuracy, but only up to a limited number of layers and neurons due to overfitting issues. In addition, the hyperparameter solution is closely related to the size of the historical database \cite{toubeau2018deep}. A deep learning model with a larger amount of hidden layers and neurons requires a large amount of data to accurately estimate the parameters. In the case study considered, there are only 157 days of data with a 15 minutes resolution. Thus, we decided to restrict the number of layers and neurons to select a smaller model that performs better with the available information.

\section{Probabilistic forecasting quality assessment}\label{sec:scoring}

For predictions in any form, one must differentiate between their quality and their value \cite{morales2013integrating}. Forecast quality corresponds to the ability of the forecasts to genuinely inform of future events by mimicking the characteristics of the processes involved. Forecast value relates, instead, to the benefits from using forecasts in a decision-making process such as participation in the electricity market. This section proposes quality metrics based on the framework proposed by \cite{lauret2019verification}. The value assessment is not in the scope of this paper as it would require to consider a decision-making process.

\subsection{Continuous rank probability score (CRPS)}

A score is said to be proper if it ensures that the perfect forecasts should be given the best score value \cite{gneiting2007strictly}. It is the case of the Continuous Rank Probability Score (CRPS) that penalizes the lack  of resolution of the predictive distributions as well as biased forecasts. For deterministic forecasts, the CRPS turns out to be the Mean Absolute Error (MAE). The energy form \cite{gneiting2007strictly} of the CRPS (NRG) is selected in this study
\begin{align}\label{eq:CRPS_NRG}
\text{CRPS}^\text{NRG}_{t,k}(\hat{F}_{t+k|t}, y_{t+k}) = & \mathbb{E}_{\hat{F}_{t+k|t}} |X-y_{t+k}| \notag \\
& - \frac{1}{2} \mathbb{E}_{\hat{F}_{t+k|t}} |X-X'|,
\end{align}
where $X$ and $X'$ are two independent copies of a random variable with distribution function $\hat{F}_{t+k|t}$ and finite first moment.
A CRPS estimator (eNRG) of the energy form is provided by \cite{zamo2018estimation} when the CDF is only known at $t+k$ through a $Q$-ensemble of quantile forecasts $ \mathcal{Q}_{t,k} = \{\hat{y}^{(q)}_{t+k|t} \}_{q \in \mathcal{Q}}$
\begin{align}\label{eq:CRPS_eNRG}
\text{CRPS}^\text{eNRG}_{t,k}(\mathcal{Q}_{t,k}, y_{t+k}) & = \frac{1}{Q}\sum_{q \in \mathcal{Q}}|\hat{y}^{(q)}_{t+k|t}-y_{t+k}| \notag \\ 
& - \frac{1}{2 Q^2} \sum_{q \in \mathcal{Q}} \sum_{q' \in\mathcal{Q}} |\hat{y}^{(q)}_{t+k|t} - \hat{y}^{(q')}_{t+k|t}|.
\end{align}
Finally, $\text{CRPS}^\text{eNRG}(k)$ is the mean over the evaluation set for a given forecasting time period $k$, and $\text{CRPS}^\text{eNRG}$ is the average over all $k$ with $k_1 \leq k \leq k_T$.

\subsection{Interval score (IS)}

The Interval Score (IS) is a proper score proposed by \cite{gneiting2007strictly} to specifically assess the quality of central $(1-\alpha)$ prediction interval forecasts. The IS rewards narrow prediction intervals but penalizes, with the penalty term that depends on $\alpha$, the forecasts for which the observation is outside the interval. The averaged IS over an evaluation set of length $M$ and $\forall k, k_1 \leq k \leq k_T$ is
\begin{align}\label{eq:IS_def}
\text{IS}_\alpha & = \frac{1}{M} \sum_{t=1}^{M} \frac{1}{T}  \sum_{k=k_1}^{k_T}   (\hat{y}^{(1-\alpha/2)}_{t+k|t} - \hat{y}^{(\alpha/2)}_{t+k|t}) \notag \\
& + \frac{2}{\alpha} (\hat{y}^{(\alpha/2)}_{t+k|t} - y_{t+k}) \boldsymbol{1}(y_{t+k} \leq \hat{y}^{(\alpha/2)}_{t+k|t} ) \notag \\
& + \frac{2}{\alpha} ( y_{t+k} - \hat{y}^{(1-\alpha/2)}_{t+k|t}) \boldsymbol{1}(y_{t+k} \geq \hat{y}^{(1-\alpha/2)}_{t+k|t} ) .
\end{align}

\section{The ULi\`ege case study}\label{sec:case_study}

\subsection{Case study description}

The ULi\`ege case study is composed of a PV generation plant with an installed capacity of 466.4 kW. The PV generation has been monitored on a minute basis from $\startdataset$ to $\lastdataset$, 157 days in total, and the data is resampled to 15 minutes. The set of quantiles is $\mathcal{Q} =  \{0.1, 0.2, \ldots, 0.9\}$ for both the day-ahead and intraday forecasts. 
Numerical experiments are performed on an Intel Core i7-8700 3.20 GHz based computer with 12 physical CPU cores and 32 GB of RAM running on Ubuntu 18.04 LTS. 

\subsection{Numerical settings}

The MAR regional climate model \cite{fettweis2017reconstructions} provided by the Laboratory of Climatology of the Li\`ege University is forced by GFS (Global Forecast System) to compute weather forecasts on a six hours basis, four-time gates per day at 00:00, 06:00, 12:00, and 18:00 with a 10 day horizon and a 15 minutes resolution. The solar irradiance and air temperature at 2 meters are normalized by a standard scaler and used as inputs to the forecasting models.

A $k$-fold cross-validation is strategy is used to compute average scores over several testing sets to mitigate the dependency of the results to specific days of the dataset. The dataset is divided into $k$ parts of equal length, and there are $k$ possible testing sets $1 \leq i \leq k$. For a given testing set $i$, the models are trained over the $k-1$ parts of the dataset. Eleven pairs of fixed lengths of 142 and 15 days are built. One pair is used to conduct the hyperparameters sensitivity analysis, and the ten others for testing where the scores are averaged.
The Mean Absolute Error (NMAE) and Root Mean Squared Error (NRMSE) are introduced to evaluate the point forecasts. The MAE, RMSE, CRPS, and IS are normalized by the PV total installed capacity with NMAE and NRMSE the normalized MAE and RMSE.

The day-ahead models, MLP, LSTM, and GBR compute forecasts at 12:00 for the next day.
Four intraday time gates are considered at 00:00, 06:00, 12:00, and 18:00. The intraday forecasts of time gate 00:00 are computed by the day-ahead models using only the weather forecasts. Then, the next three intraday forecasts are computed by intraday models where the MLP, ED-1, and ED-2, models use the weather forecasts and the last three hours of PV generation.

The day-ahead and the first intraday predictions are delivered for the 96 quarters of the next day from 00:00 to 23:45 indexed by time steps $0 \leq k \leq 95$. The prediction horizons span from 12 to 36 hours, for the day-ahead gate 12:00, and 0 to 24 hours, for the intraday gate 00:00.
The prediction horizon is cropped to $11 \leq k \leq 80$ because the PV generation is always 0 for time steps $0 \leq k \leq 10$ and $81 \leq k \leq 95$ on the ULi\`ege case study.
The next three intraday predictions are performed for the 72, 48, and 24 next quarters of the day corresponding to the gates 06:00, 12:00, and 18:00. Therefore, the prediction horizons span from 0 to 18 hours, 0 to 12 hours, and 0 to 6 hours. The intraday forecasting time periods are $24 \leq k \leq 80$, $48 \leq k \leq 80$, and $72 \leq k \leq 80$.
Table \ref{tab:tcpu_comparison} compares the mean and the standard deviation of the computation times, over the ten learning sets, to train the point and quantile forecast models\footnote{The day-ahead and intraday LSTM training times are identicals for both point and quantile forecasts as they only take the weather forecasts as inputs.}.
\begin{table}[htbp]
\renewcommand{\arraystretch}{1.25}
	\begin{center}
		\begin{tabular}{lrrr}
			\hline  \hline
			day-ahead &  MLP & LSTM & GBR \\ \hline
			point    & 5.3  (0.1)  & 23.7  (0.3) & 3.4 (0.1)  \\
			quantile & 7.6  (0.2)  & 69.0  (0.6) & 44.6 (0.4)  \\ \hline 
			
			intraday  &  MLP & ED-1  & ED-2 \\ \hline
			point     & 5.0  (0.1)   & 5.2  (0.1) & 17.2 (0.2) \\
			quantile  & 17.9  (0.2)  & 6.4  (0.2) & 18.0 (0.3) \\ \hline \hline
		\end{tabular}
		\caption{Training computation time (s).}
		\label{tab:tcpu_comparison}
	\end{center}
\end{table}

\subsection{Day-ahead results}\label{sec:dad_results}

\begin{figure}[htbp]
	\centering
	\begin{subfigure}{.25\textwidth}
		\centering
		\includegraphics[width=\linewidth]{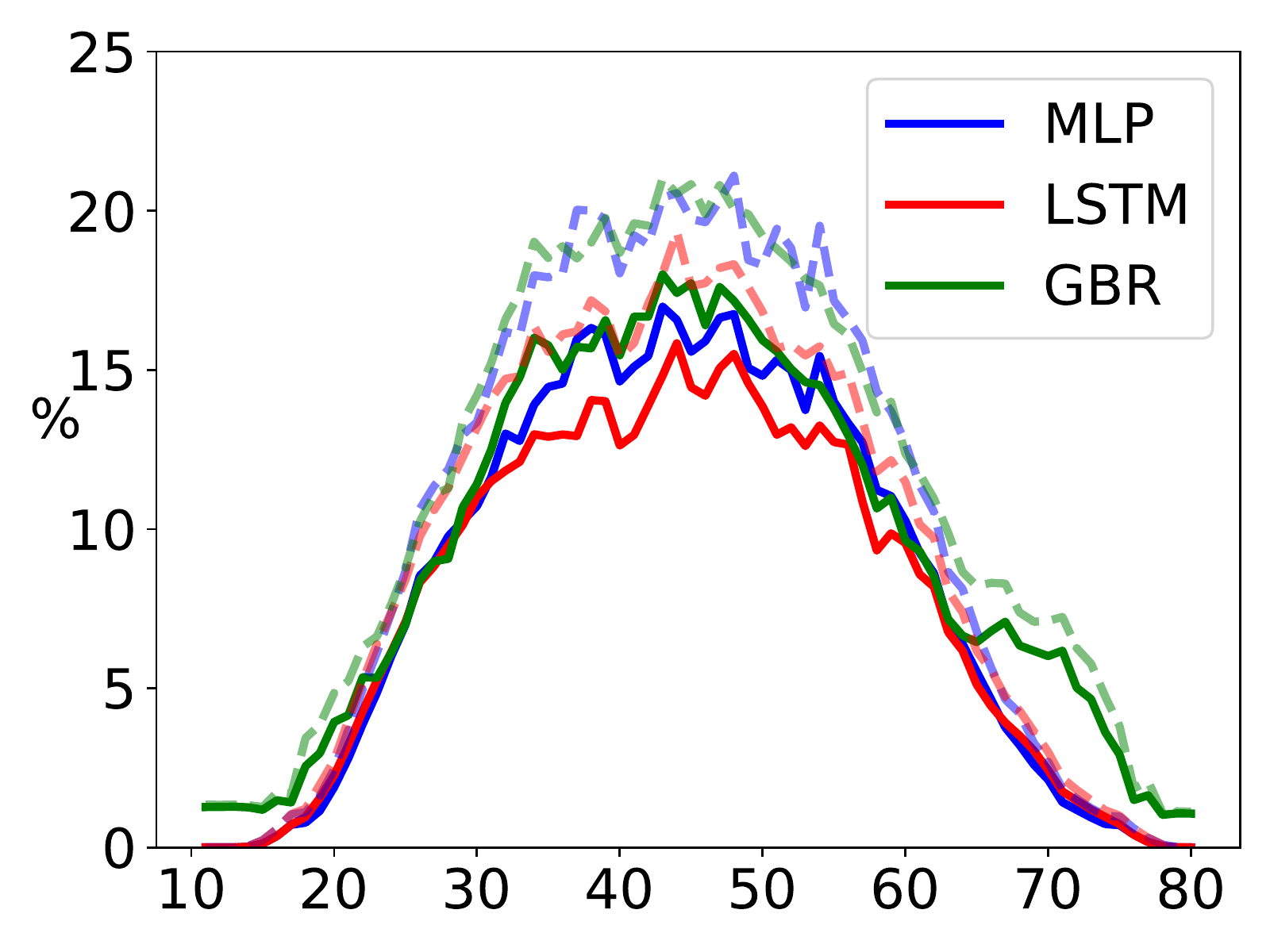}
		\caption{NMAE and NRMSE.}
		\label{fig:dad_point_comparison}
	\end{subfigure}%
	\begin{subfigure}{.25\textwidth}
		\centering
		\includegraphics[width=\linewidth]{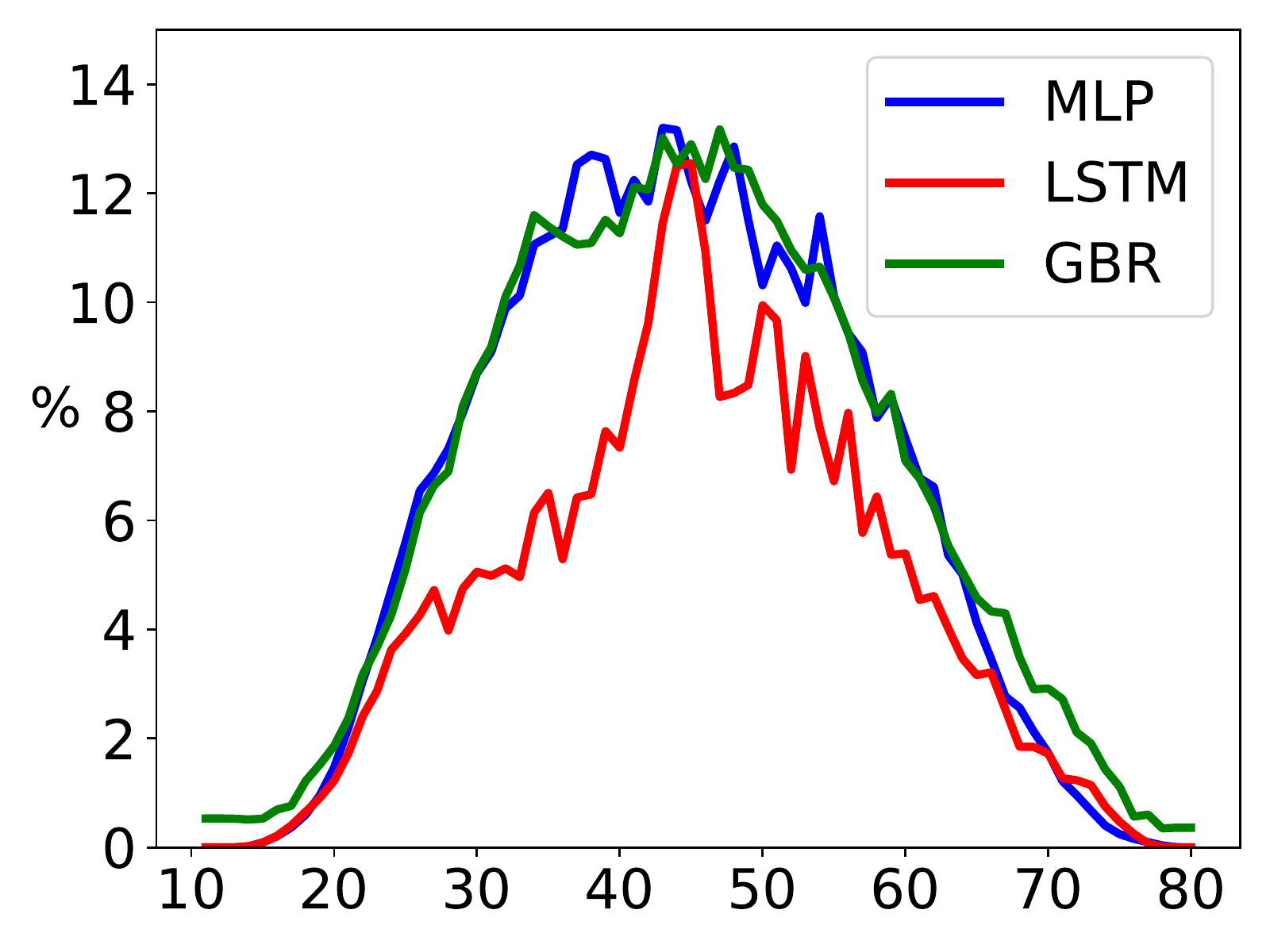}
		\caption{CRPS.}	
		\label{fig:dad_quantile_comparison}
	\end{subfigure}
	\caption{Day-ahead models results with the NMAE (plain lines), NRMSE (dashed lines), and CRPS.}
	\label{fig:dad_comparison}
\end{figure}

Figure \ref{fig:dad_point_comparison} compares the NMAE (plain lines), NRMSE (dashed lines), and Figure \ref{fig:dad_quantile_comparison} the CRPS per forecasting time periods $k$ of the day-ahead models of gate 12:00. Table \ref{tab:dad_comparison} provides the mean and standard deviation of the NMAE, NRMSE, and CRPS. The LSTM achieved the best results for both point and quantile forecasts. Figures \ref{fig:forecasts_plot_MLP_dad}, \ref{fig:forecasts_plot_LSTM_dad}, and \ref{fig:forecasts_plot_GBR_dad} compare the MLP, LSTM, and GBR day-ahead quantile and point forecasts (black line named dad 12) of gate 12:00 on $\figuredatecomparison$ with the observation in red. One can see that the predicted intervals of the LSTM model better encompass the actual realizations of uncertainties than the MLP and GBR.
\begin{table}[htbp]
\renewcommand{\arraystretch}{1.25}
	\begin{center}
		\begin{tabular}{llrrr}
			\hline  \hline
			Score & Gate &  MLP & LSTM & GBR\\
			\hline
			\multirow{2}{*}{NMAE} & 12 & 8.2 (1.2) & 7.6  (1.5) &  9.2 (0.9)\\
			 & 24 & 7.9 (1.2) & 7.7  (1.6) &  9.0 (0.8)\\ \hline  
			
			\multirow{2}{*}{NRMSE} & 12 & 10.2  (1.4) & 9.2  (1.6) & 11.2 (0.9) \\
			 & 24 & 9.7   (1.2) & 9.4  (1.8) & 10.9 (0.8) \\ \hline 
			
			\multirow{2}{*}{CRPS} & 12 & 6.2  (1.1) & 4.4  (0.2) & 6.4 (0.7)  \\
			 & 24 & 6.2  (1.0) & 4.4  (0.2) & 6.3 (0.6)   \\ \hline	 \hline		
		\end{tabular}
		\caption{Day-ahead models results.}
		\label{tab:dad_comparison}
	\end{center}
\end{table}

\subsection{Intraday results}\label{sec:intra_results}

\begin{figure}[tb]
	\centering
	\begin{subfigure}{.25\textwidth}
		\centering
		\includegraphics[width=\linewidth]{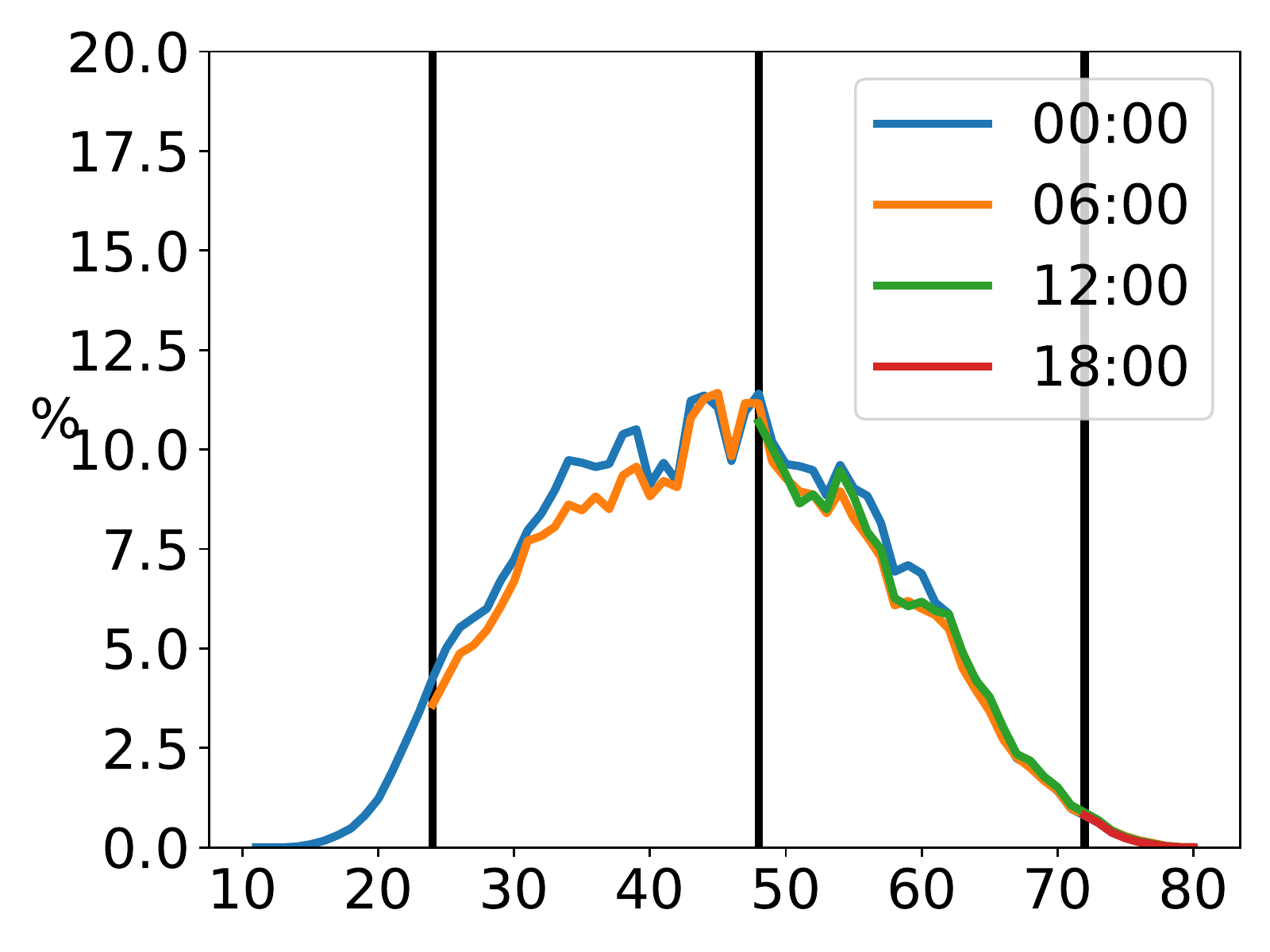}
		\caption{ED-1.}
	\end{subfigure}%
	\begin{subfigure}{.25\textwidth}
		\centering
		\includegraphics[width=\linewidth]{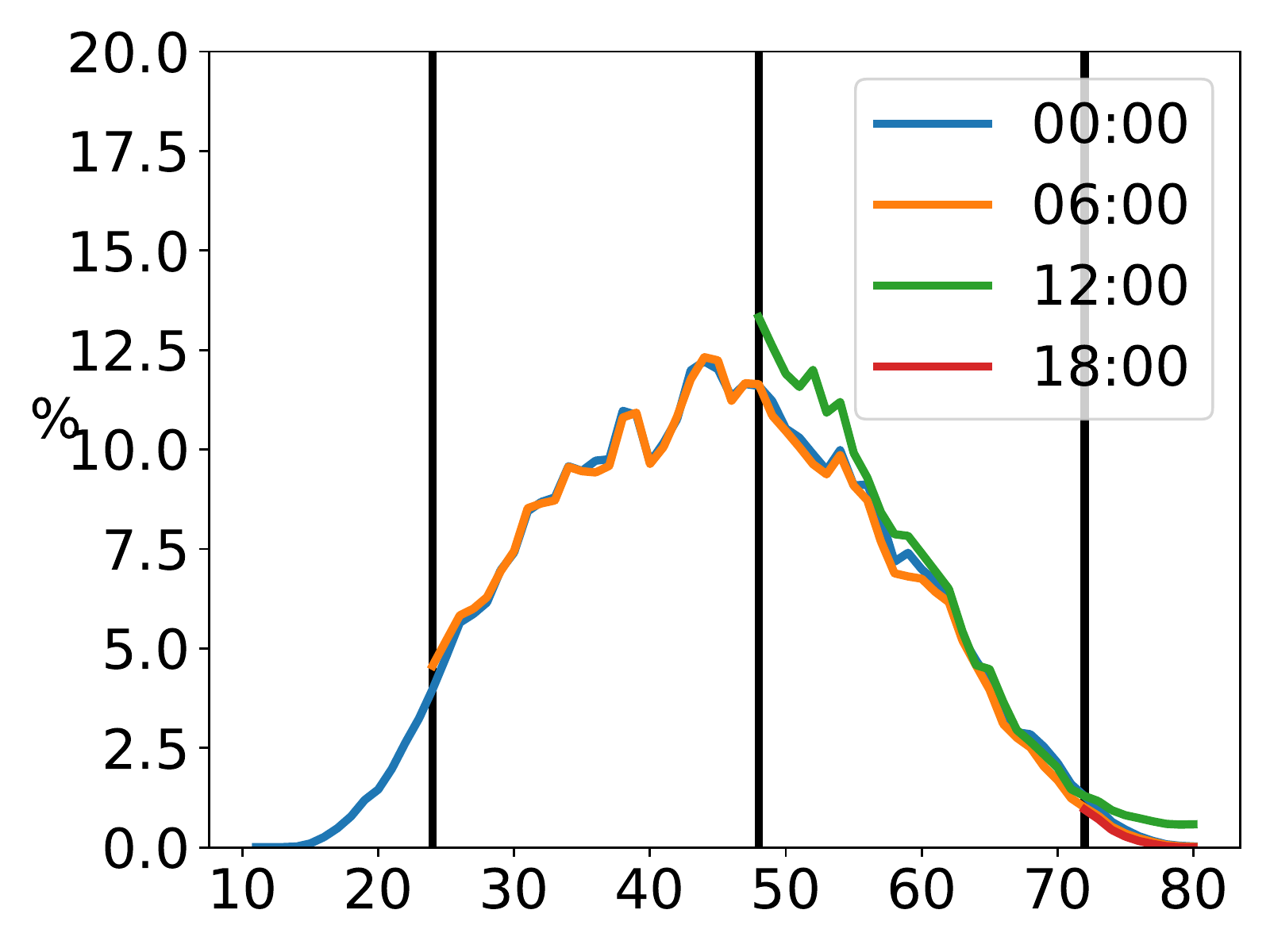}
		\caption{ED-2.}
	\end{subfigure}
\begin{subfigure}{.25\textwidth}
	\centering
	\includegraphics[width=\linewidth]{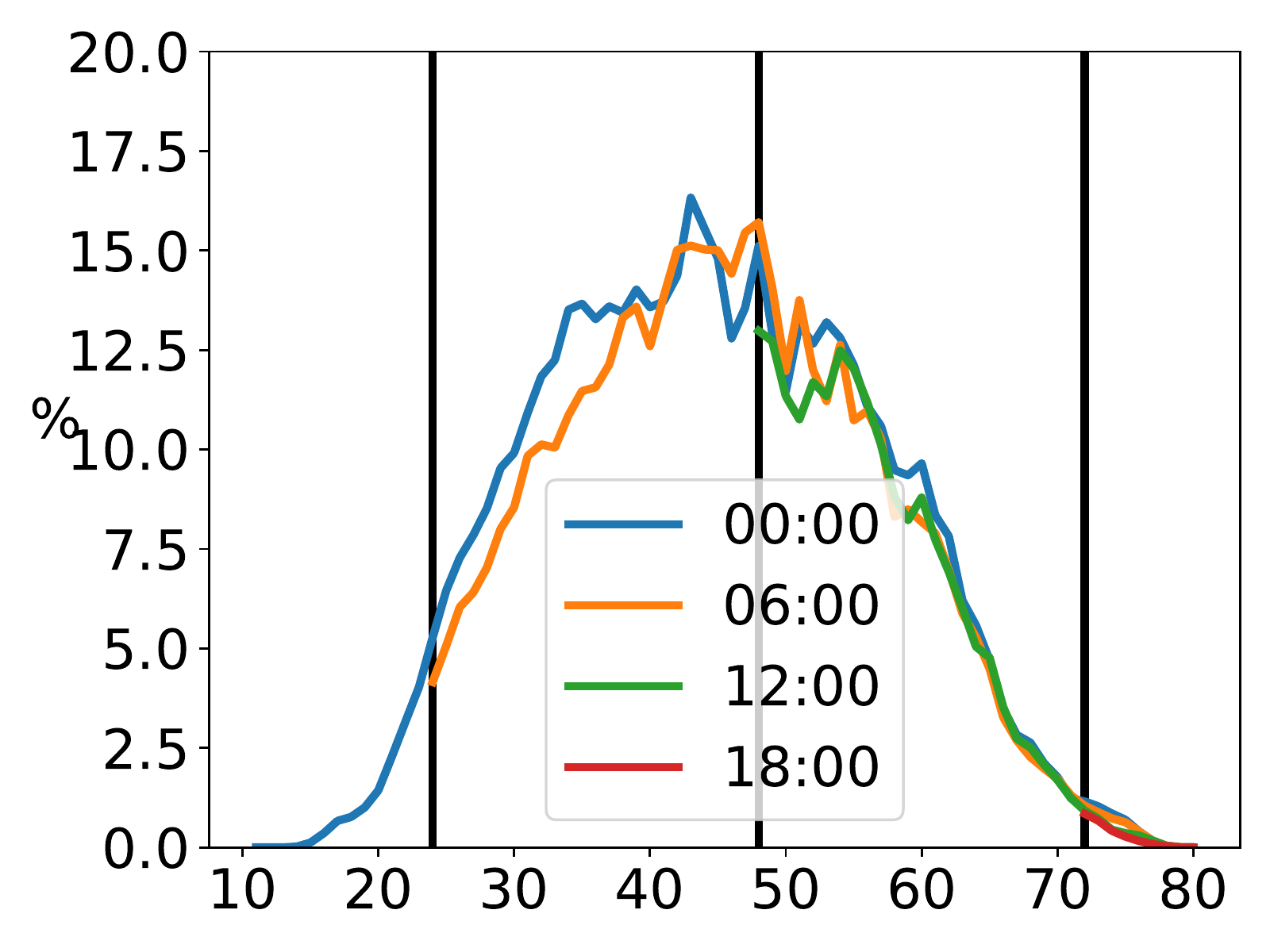}
	\caption{MLP.}
\end{subfigure}%
\begin{subfigure}{.25\textwidth}
	\centering
	\includegraphics[width=\linewidth]{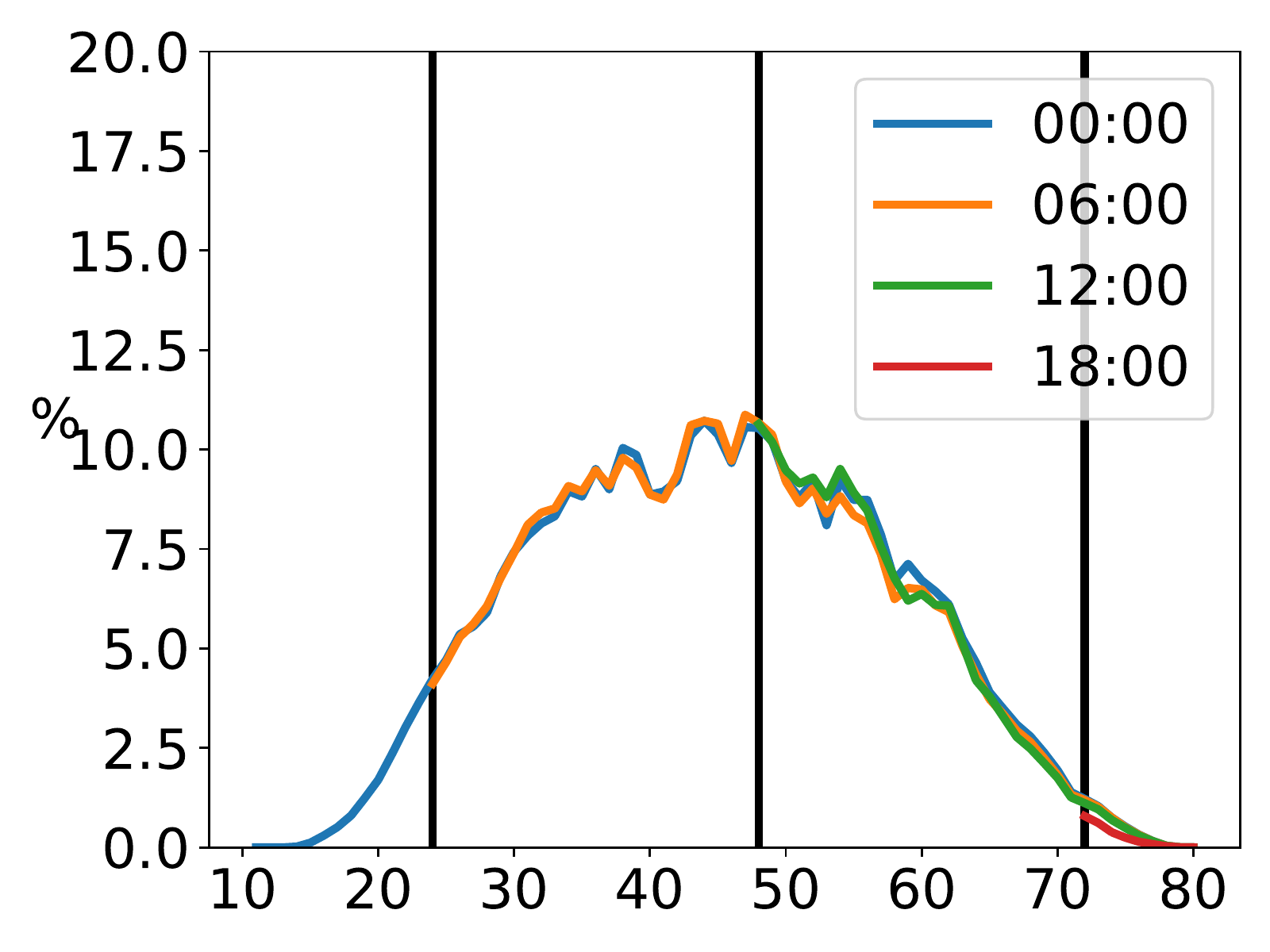}
	\caption{LSTM.}
\end{subfigure}
	\caption{Intraday models CRPS results.}
	\label{fig:intra_CRPS}
\end{figure}

\begin{figure}[htbp]
	\centering
	\begin{subfigure}{.25\textwidth}
		\centering
		\includegraphics[width=\linewidth]{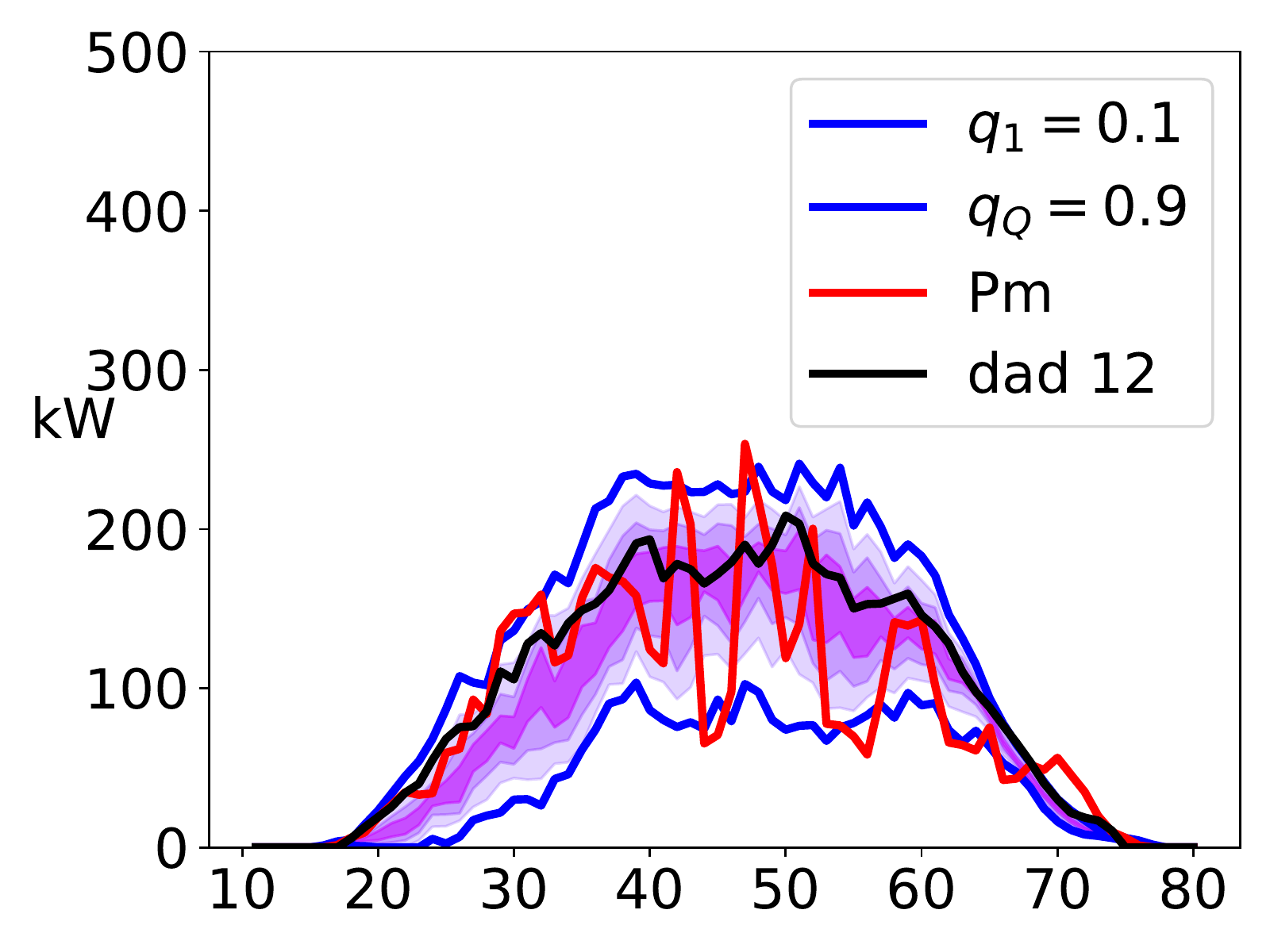}
		\caption{MLP day-ahead.}
		\label{fig:forecasts_plot_MLP_dad}
	\end{subfigure}%
	\begin{subfigure}{.25\textwidth}
	\centering
	\includegraphics[width=\linewidth]{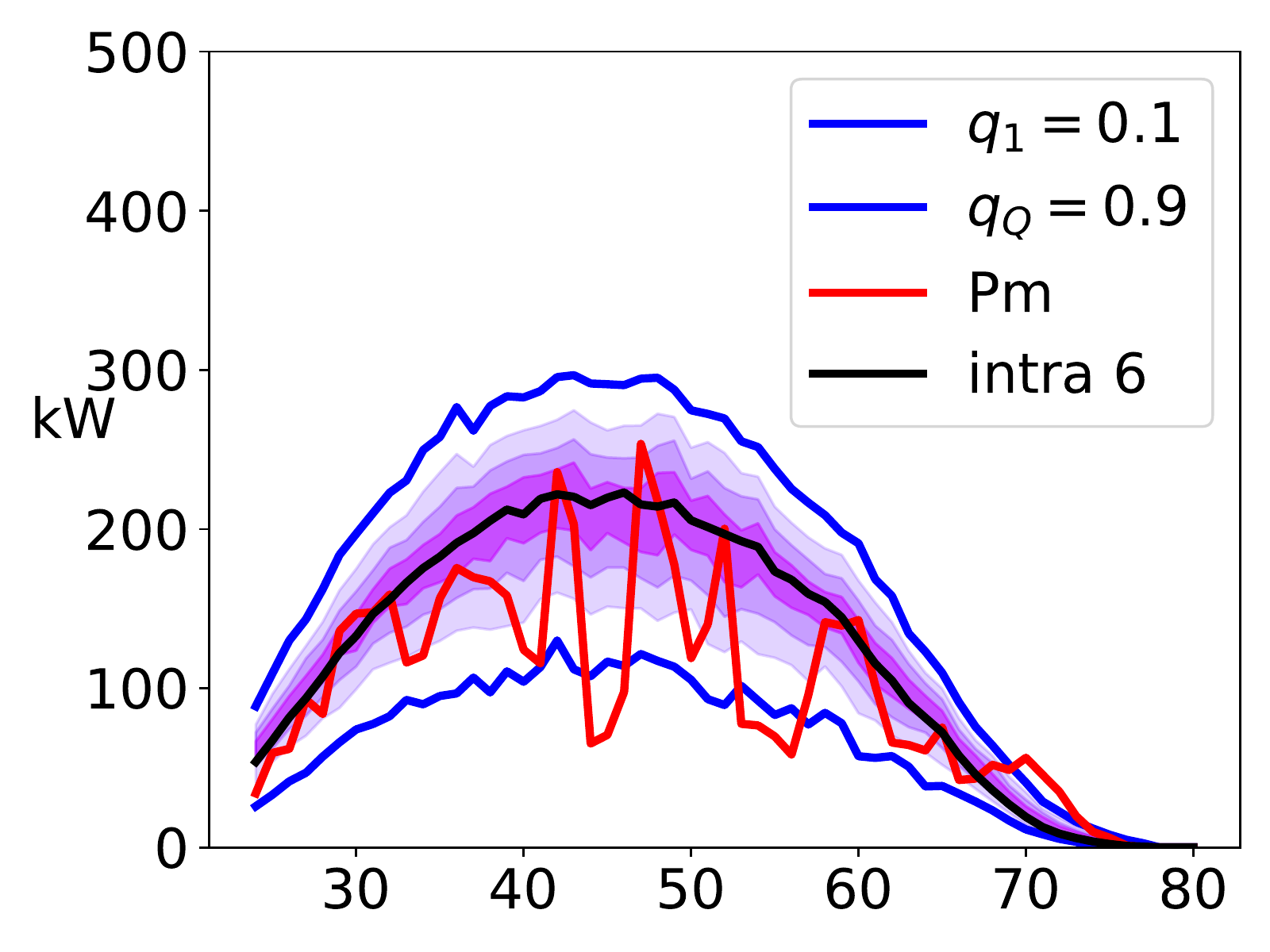}
	\caption{ED-1 intraday.}
	\label{fig:forecasts_plot_ED1_intra}
\end{subfigure}
	\begin{subfigure}{.25\textwidth}
		\centering
		\includegraphics[width=\linewidth]{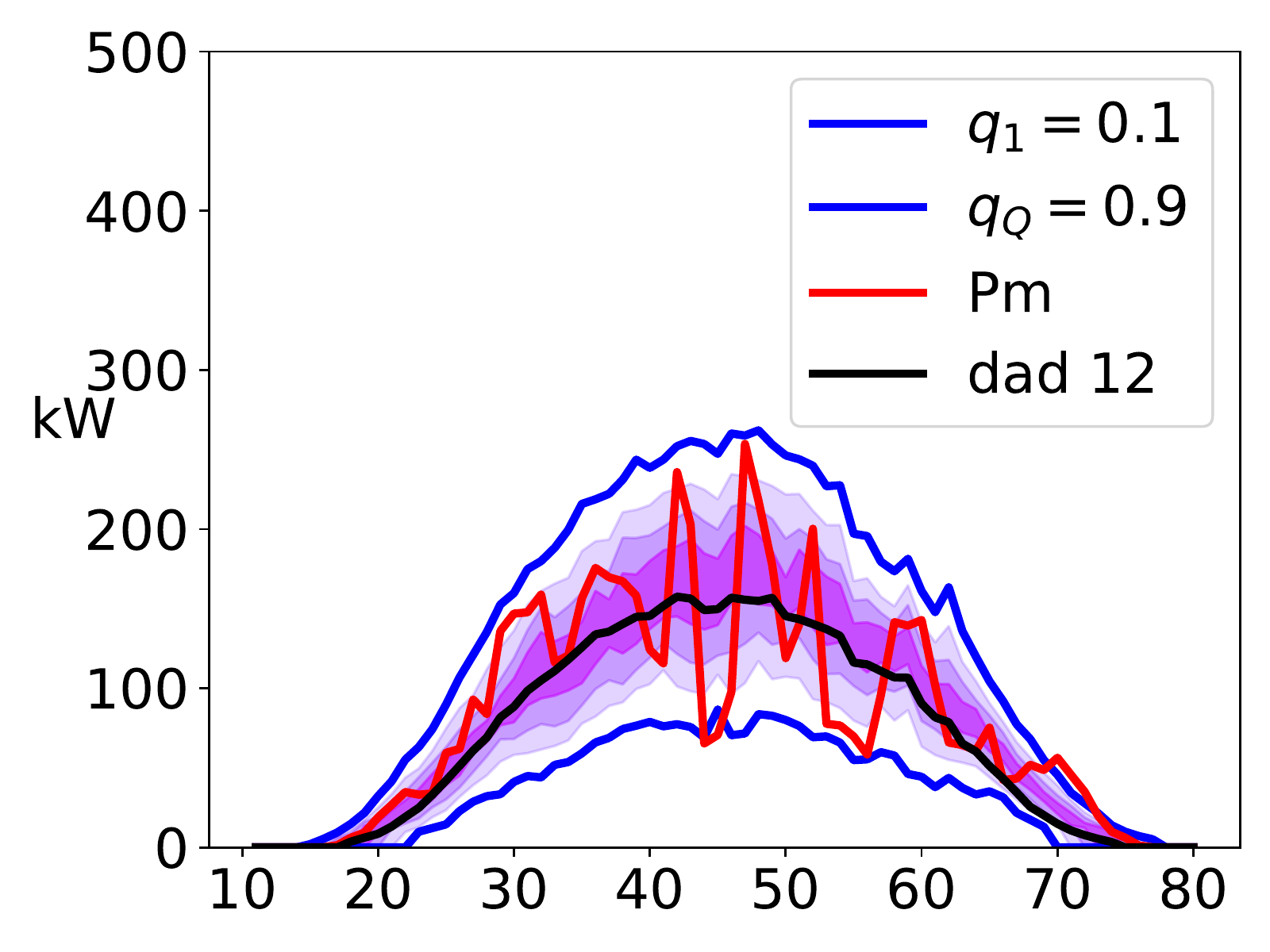}
		\caption{LSTM day-ahead.}
		\label{fig:forecasts_plot_LSTM_dad}
	\end{subfigure}%
	\begin{subfigure}{.25\textwidth}
	\centering
	\includegraphics[width=\linewidth]{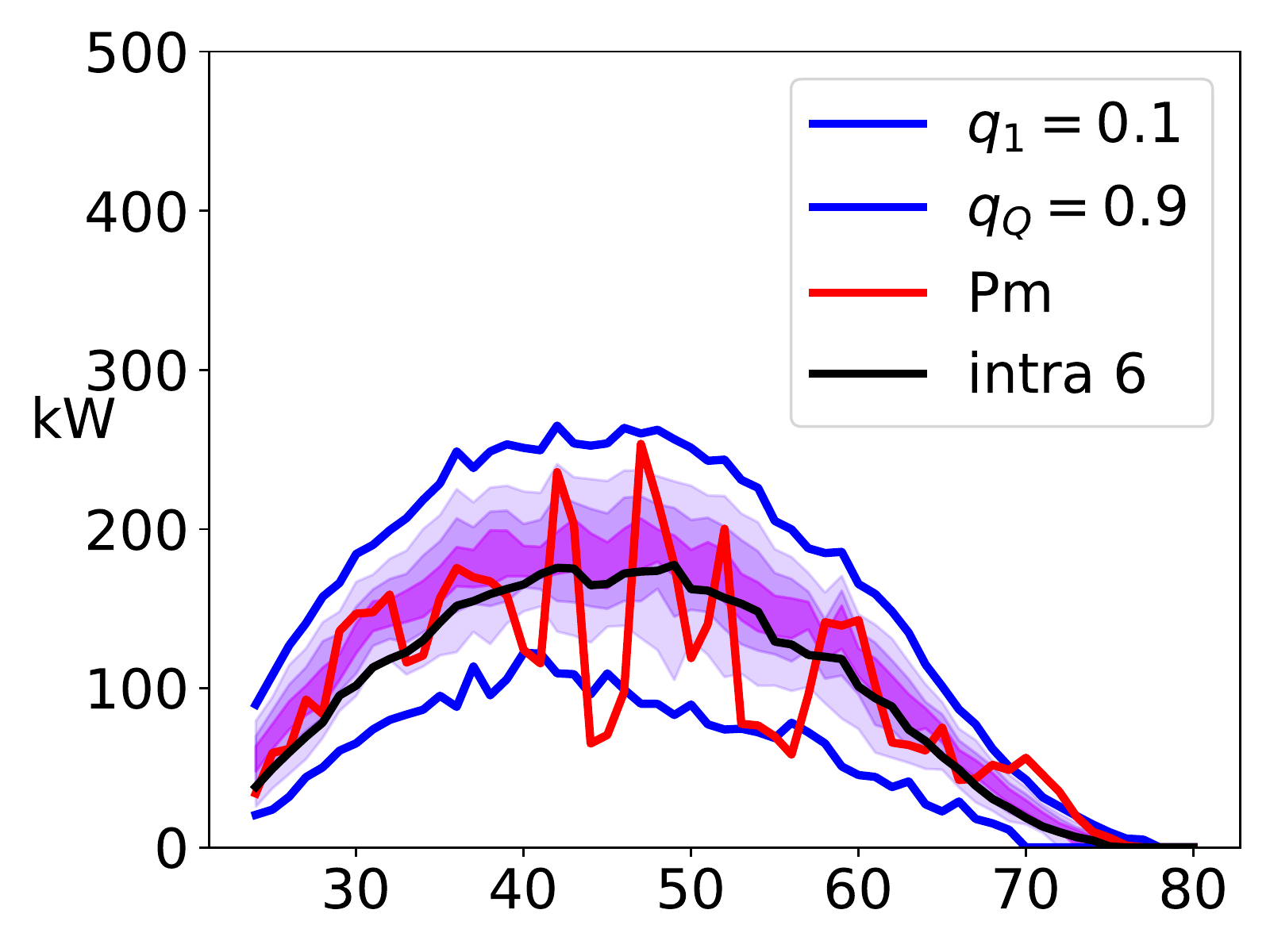}
	\caption{LSTM intraday.}
	\label{fig:forecasts_plot_LSTM_intra}
\end{subfigure}
	\begin{subfigure}{.25\textwidth}
		\centering
		\includegraphics[width=\linewidth]{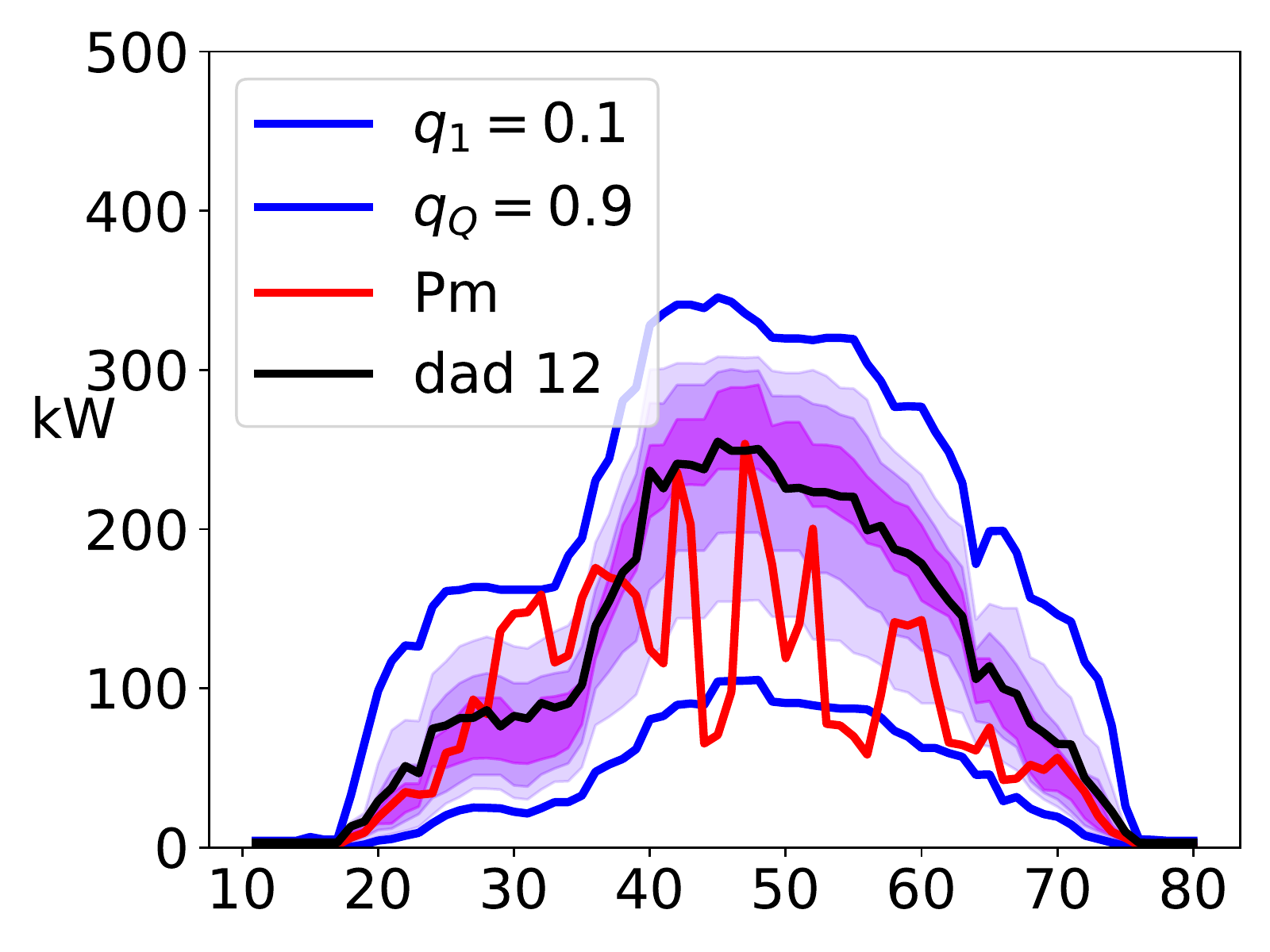}
		\caption{GBR day-ahead.}
		\label{fig:forecasts_plot_GBR_dad}
	\end{subfigure}%
	\begin{subfigure}{.25\textwidth}
		\centering
		\includegraphics[width=\linewidth]{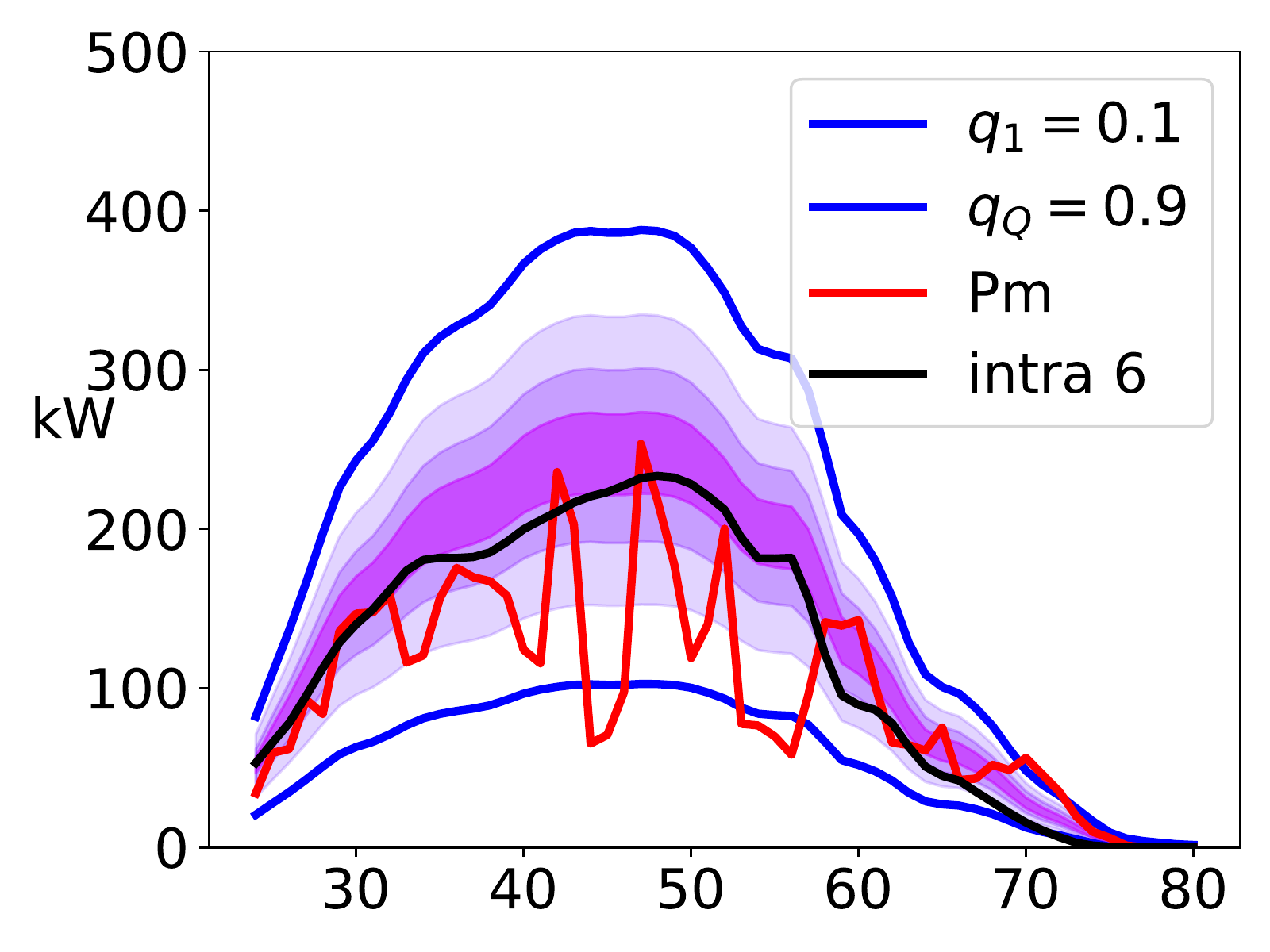}
		\caption{ED-2 intraday.}
		\label{fig:forecasts_plot_ED2_intra}
	\end{subfigure}
	\caption{Quantiles vs point forecasts of day-ahead models of gate 12:00 (left), and intraday models of gate 06:00 (right) on $\figuredatecomparison$, the observations are in red.}
	\label{fig:forecasts_plot}
\end{figure}

Table \ref{tab:intra_scores} provides the averaged NMAE, NRMSE, and CRPS per gate of intraday models. The LSTM achieved the best NMAE and NRMSE for the 06:00 gate and the ED-1 achieved the best NMAE and NRMSE for the 12:00 gate and the best CRPS for both gates. Figure \ref{fig:intra_CRPS} compares the CRPS per forecasting time periods $k$ of the intraday models. The ED-1 benefits from the last PV generation observations. Indeed, some CRPS values for both 06:00 and 12:00 gates are below the ones of 00:00 gate.
Table \ref{tab:intra_WS} provides the Interval score of intraday models for 80 \%, 60 \%, 40 \%, and 20 \% width of central intervals. The ED-1 model achieved the best results for both 06:00 and 12:00 gates and all prediction intervals except for the 06:00 gate and the prediction interval width of 80 \% where it is ED-2. The LSTM achieved close results to the ED-1. 
Figures \ref{fig:forecasts_plot_ED1_intra}, \ref{fig:forecasts_plot_LSTM_intra}, and \ref{fig:forecasts_plot_ED2_intra} compare the ED-1, LSTM, and ED-2 intraday quantile and point forecasts (black line named intra 6) of 06:00 gate on $\figuredatecomparison$ with the observation in red. Generally, one can see that the predicted intervals of ED-1 and LSTM models better encompass the actual realizations of uncertainties than ED-2.
\begin{table}[!htb]
\renewcommand{\arraystretch}{1.25}
	\begin{center}
		\begin{tabular}{llrrrr}
			\hline  \hline
			Score & Gate &  MLP & ED-1 & ED-2 & LSTM\\
			\hline
			\multirow{2}{*}{NMAE} & 6  & 8.9  (1.0)  & 8.5  (1.4)  &9.4  (1.0 ) & 7.6  (1.5) \\
			 & 12 & 6.7  (1.4) & 6.4  (1.3 ) & 7.1  (1.1)  & 7.2  (1.1)\\ \hline 
			
			\multirow{2}{*}{NRMSE} &6  & 10.9  (0.9) & 10.3   (1.3)   & 11.3  (1.1)  & 7.7  (1.6) \\
			 &12 & 8.7   (1.3) & 7.8    (1.2)   & 8.5   (1.2)  & 9.4  (1.8)\\ \hline 
			
			\multirow{2}{*}{CRPS} &6  & 8.1  (0.7) & 5.9  (0.9) & 6.6  (0.7) & 6.2  (0.7) \\
			 &12 & 5.8  (1.2) & 4.5  (0.7) & 5.6  (1.8) & 4.7  (0.5) \\ \hline \hline
			
		\end{tabular}
		\caption{Intraday models NMAE, NRMSE, CRPS results.}
		\label{tab:intra_scores}
	\end{center}
\end{table}
\begin{table}[!htb]
\renewcommand{\arraystretch}{1.25}
	\begin{center}
		\begin{tabular}{llrrrr}
			\hline  \hline
			Width & Gate & MLP & ED-1 & ED-2 & LSTM \\
			\hline
			\multirow{2}{*}{80 \%} & 6  & 24.4  (2.9) &  14.9  (4.0) & 13.9  (4.9) & 19.3 (4.2)\\
			 & 12 & 17.4  (3.5) &  10.6  (1.8) & 11.6  (10.1) & 9.6 (2.0)\\ \hline  
			\multirow{2}{*}{60 \%} & 6  & 37.6  (3.2) &  29.9  (5.0) & 32.2  (4.2) & 30.7 (4.6)\\
			 & 12 & 27.2  (4.3) &  22.4  (4.2) & 27.5  (10.8)&  22.6 (3.1)\\ \hline  
			\multirow{2}{*}{40 \%} & 6  & 58.0  (4.5) &  50.1  (6.5) & 57.2  (6.0) & 51.6 (5.8)\\
			& 12 & 42.4  (6.9)  & 37.7  (5.9) & 48.1  (16.8) & 39.2 (4.9)\\ \hline  
			\multirow{2}{*}{20 \%} & 6  & 111.8  (8.4) & 97.1  (11.7) & 112.1  (10.3)& 99.5 (10.4)\\
			& 12 & 81.5  (13.8) &  72.7  (10.0) & 94.8  (32.1) & 76.5 (8.0)\\ \hline  \hline
		\end{tabular}
		\caption{Intraday models IS results.}
		\label{tab:intra_WS}
	\end{center}
\end{table}

\section{Conclusion}\label{sec:conclusion}

An encoder-decoder architecture is implemented on the intraday scale to produce accurate forecasts. It efficiently captures the contextual information composed of past PV observations and future weather forecasts, while capturing the temporal dependency between forecasting time periods over the entire forecasting horizon.  The models are compared by using a $k$-fold cross-validation methodology and quality metrics on a real case study composed of the PV generation of the parking rooftops of the Li\`ege University.
The best day-ahead model for both point and quantile forecasts is a neural network composed of a LSTM cell and an additional feed-forward layer. 
Then, the encoder-architecture composed of a LSTM-MLP yields accurate and calibrated forecast distributions learned from the historical dataset in comparison with the MLP and LSTM-LSTM models for the intraday point and quantile forecasts. However, the LSTM produced similar results.
%
Several extensions are under investigation. First, considering a larger dataset of at least one full year to take into account the entire PV seasonality. Second, developing a PV scenario approach based on the encoder-decoder architecture. 

\bibliographystyle{ieeetr}
\bibliography{biblio}

\begin{thebibliography}{10}

\bibitem{morales2013integrating}
J.~M. Morales, A.~J. Conejo, H.~Madsen, P.~Pinson, and M.~Zugno, {\em
  Integrating renewables in electricity markets: operational problems},
  vol.~205.
\newblock Springer Science \& Business Media, 2013.

\bibitem{koenker1978regression}
R.~Koenker and G.~Bassett~Jr, ``Regression quantiles,'' {\em Econometrica:
  journal of the Econometric Society}, pp.~33--50, 1978.

\bibitem{hong2016probabilistic}
T.~Hong, P.~Pinson, S.~Fan, H.~Zareipour, A.~Troccoli, and R.~J. Hyndman,
  ``Probabilistic energy forecasting: Global energy forecasting competition
  2014 and beyond,'' 2016.

\bibitem{golestaneh2016very}
F.~Golestaneh, P.~Pinson, and H.~B. Gooi, ``Very short-term nonparametric
  probabilistic forecasting of renewable energy generation—with application
  to solar energy,'' {\em IEEE Transactions on Power Systems}, vol.~31, no.~5,
  pp.~3850--3863, 2016.

\bibitem{toubeau2018deep}
J.-F. Toubeau, J.~Bottieau, F.~Vall{\'e}e, and Z.~De~Gr{\`e}ve, ``Deep
  learning-based multivariate probabilistic forecasting for short-term
  scheduling in power markets,'' {\em IEEE Transactions on Power Systems},
  vol.~34, no.~2, pp.~1203--1215, 2018.

\bibitem{bottieau2019very}
J.~Bottieau, L.~Hubert, Z.~De~Gr{\`e}ve, F.~Vall{\'e}e, and J.-F. Toubeau,
  ``Very-short-term probabilistic forecasting for a risk-aware participation in
  the single price imbalance settlement,'' {\em IEEE Transactions on Power
  Systems}, vol.~35, no.~2, pp.~1218--1230, 2019.

\bibitem{lauret2017probabilistic}
P.~Lauret, M.~David, and H.~T. Pedro, ``Probabilistic solar forecasting using
  quantile regression models,'' {\em energies}, vol.~10, no.~10, p.~1591, 2017.

\bibitem{lauret2019verification}
P.~Lauret, M.~David, and P.~Pinson, ``Verification of solar irradiance
  probabilistic forecasts,'' {\em Solar Energy}, vol.~194, pp.~254--271, 2019.

\bibitem{pinson2007non}
P.~Pinson, H.~A. Nielsen, J.~K. M{\o}ller, H.~Madsen, and G.~N. Kariniotakis,
  ``Non-parametric probabilistic forecasts of wind power: required properties
  and evaluation,'' {\em Wind Energy: An International Journal for Progress and
  Applications in Wind Power Conversion Technology}, vol.~10, no.~6,
  pp.~497--516, 2007.

\bibitem{dumas2020stochastic}
J.~Dumas, B.~Corn{\'e}lusse, A.~Giannitrapani, S.~Paoletti, and A.~Vicino,
  ``Stochastic and deterministic formulations for capacity firming
  nominations,'' in {\em 2020 International Conference on Probabilistic Methods
  Applied to Power Systems (PMAPS)}, pp.~1--7, IEEE, 2020.

\bibitem{fettweis2017reconstructions}
X.~Fettweis, J.~Box, C.~Agosta, C.~Amory, C.~Kittel, C.~Lang, D.~van As,
  H.~Machguth, and H.~Gall{\'e}e, ``Reconstructions of the 1900--2015 greenland
  ice sheet surface mass balance using the regional climate {MAR} model,'' {\em
  Cryosphere (The)}, vol.~11, pp.~1015--1033, 2017.

\bibitem{hastie2009elements}
T.~Hastie, R.~Tibshirani, and J.~Friedman, {\em The elements of statistical
  learning: data mining, inference, and prediction}.
\newblock Springer Science \& Business Media, 2009.

\bibitem{scikit-learn}
F.~Pedregosa, G.~Varoquaux, A.~Gramfort, V.~Michel, B.~Thirion, O.~Grisel,
  M.~Blondel, P.~Prettenhofer, R.~Weiss, V.~Dubourg, {\em et~al.},
  ``Scikit-learn: Machine learning in python,'' {\em the Journal of machine
  Learning research}, vol.~12, pp.~2825--2830, 2011.

\bibitem{paszke2017automatic}
A.~Paszke, S.~Gross, S.~Chintala, G.~Chanan, E.~Yang, Z.~DeVito, Z.~Lin,
  A.~Desmaison, L.~Antiga, and A.~Lerer, ``Automatic differentiation in
  pytorch,'' in {\em NIPS-W}, 2017.

\bibitem{tensorflow2015-whitepaper}
M.~Abadi, A.~Agarwal, P.~Barham, E.~Brevdo, Z.~Chen, C.~Citro, G.~S. Corrado,
  A.~Davis, J.~Dean, M.~Devin, {\em et~al.}, ``Tensorflow: Large-scale machine
  learning on heterogeneous systems,'' 2015.

\bibitem{gneiting2007strictly}
T.~Gneiting and A.~E. Raftery, ``Strictly proper scoring rules, prediction, and
  estimation,'' {\em Journal of the American statistical Association},
  vol.~102, no.~477, pp.~359--378, 2007.

\bibitem{zamo2018estimation}
M.~Zamo and P.~Naveau, ``Estimation of the continuous ranked probability score
  with limited information and applications to ensemble weather forecasts,''
  {\em Mathematical Geosciences}, vol.~50, no.~2, pp.~209--234, 2018.

\end{thebibliography}

\end{document}